\documentclass[UTF8,nofonts,10pt, oneside]{article}       
\usepackage{geometry}                        
\usepackage[whole]{bxcjkjatype} 
\usepackage[unicode]{hyperref}

\usepackage{authblk}

\usepackage{hyperref}
\hypersetup
{
    colorlinks=true,
    linkcolor=blue,
    citecolor=magenta,
    urlcolor=black,
}
                       
\usepackage[parfill]{parskip}            
\usepackage{graphicx,caption,subcaption,array,rotating,multirow,bigdelim}                
\usepackage{amssymb,amsthm,amsmath,mathtools}
\usepackage{mathrsfs}
\usepackage{enumitem}
\usepackage{hhline}

\let \epsilon \varepsilon

\usepackage{tikz}
\usepackage[ruled,vlined]{algorithm2e}
\usepackage{cancel}
\usepackage{verbatim}

\usepackage{caption}
\newcommand{\gyr}{
\text{gyr}
}

\newcommand{\proj}{
\text{proj}
}

\newcommand{\code}[1]{
\texttt{#1}
}

\title{\textbf{Hyperbolic Deep Learning for Chinese Natural Language Understanding}}
\author[1]{Marko Valentin Micic}
\author[2]{Hugo Chu}
\affil[ ]{Kami Intelligence Ltd, HK SAR}
\affil[1]{\href{mailto:marko.micic@kami.ai}{marko.micic@kami.ai}}
\affil[2]{\href{mailto:hugo.chu@kami.ai}{hugo.chu@kami.ai}}
\date{}

\begin{document}
\maketitle
\setcounter{MaxMatrixCols}{24}


\begin{abstract}
\noindent Recently hyperbolic geometry has proven to be effective in building embeddings that encode 
	hierarchical and entailment information. This makes it particularly suited to modelling the complex 
	asymmetrical relationships between Chinese characters and words. In this paper we first train a 
	large scale hyperboloid skip-gram model on a Chinese corpus, then apply the character 
	embeddings to a downstream hyperbolic Transformer model derived from the principles of 
	gyrovector space for Poincare disk model. In our experiments the character-based Transformer 
	outperformed its word-based Euclidean equivalent. To the best of our knowledge, this is the first 
	time in Chinese NLP that a character-based model outperformed its word-based counterpart, 
	allowing the circumvention of the challenging and domain-dependent task of Chinese Word 
	Segmentation (CWS).
\end{abstract}


\section{Introduction}

Recent advances in deep learning have seen tremendous improvements in accuracy 
and scalability in all fields of application. Natural language processing (NLP), however, still faces 
many challenges, particularly in Chinese, where even word-segmentation proves to be a difficult and 
as-yet unscalable task.

Traditional neural network architectures employed to tackle NLP tasks are generally cast in the
standard Euclidean geometry. While this setting is adequate for most deep learning tasks, where it has 
often allowed for the achievement of superhuman performance in certain areas (e.g. AlphaGo 
\cite{AlphaGo}, ImageNet \cite{ImageNet}), its performance in NLP has yet to reach the same levels. 
Recent research indicates that some of the problems NLP faces may be much more fundamental and 
mathematical in nature. It turns out that Euclidean geometry is unsuitable for properly dealing with data 
containing many hierarchical and asymmetrical relations. For instance, arbitrary trees, which are 
hierarchical data structures, cannot be properly embedded even in infinite-dimensional Euclidean 
space, yet even 2-dimensional hyperbolic space is ``large" enough to embed arbitrary trees  
\cite{DeSa2018}. Additionally, hyperbolic geometries give rise to so-called ``gyrovector spaces" 
\cite{Ungar2005} which are hyperbolic analogues of Euclidean vector spaces. Gyrovector spaces use 
gyrovector operations such as gyrovector addition (the hyperbolic analogue of vector addition) and 
scalar multiplication. Importantly, gyrovector addition is neither commutative nor associative, giving rise 
to a new mathematical way to capture asymmetrical relations in a mathematically sound framework. 

In this paper, we will first discuss the problems inherent in Chinese Word-Segmentation (CWS). Then 
we give a primer on the background and fundamental concepts of hyperbolic geometry, and then 
describe both hyperbolic and Euclidean versions of word-embedding algorithms and the intent-
classification algorithms which use these embeddings downstream. Lastly we discuss these results 
and conclude that hyperbolic geometry is a useful research direction, particularly in Chinese NLP, since 
in our experiments it appears that not only do hyperbolic versions of our algorithms outperform 
Euclidean ones, but they also appear to be able to circumvent the problem of CWS.


\section{Chinese Word Segmentation}

The state-of-the-art Chinese Word Segmentation (CWS) algorithms are bi-directional LTSMs 
\cite{Ma2018}. CWS is an important yet challenging pre-processing step for Chinese NLP; this is 
because both characters and words carry semantic meaning, which are sometimes related, and 
sometimes unrelated. For example, related characters and words include
	\begin{align*}
		&\begin{array}{| p{2cm} | p{2cm} | p{3.5cm} |}
			\hline
			\code{Char}	&	\text{游}	&	\text{to swim (verb)}	\\
			\hline
			\code{Char}	&	\text{泳}	&	\text{swimming (noun)}	\\
			\hline
			\code{Word}	&	\text{游泳}	&	\text{to swim (noun/verb)}	\\
			\hline
		\end{array}
	\end{align*}
	\begin{align*}
		&\begin{array}{| p{2cm} | p{2cm} | p{3.5cm}  |}
			\hline
			\code{Char}	&	\text{微}	&	\text{micro}	\\
			\hline
			\code{Char}	&	\text{波}	&	\text{wave}	\\
			\hline
			\code{Char}	&	\text{爐}	&	\text{oven}	\\
			\hline
			\code{Word}	&	\text{微波爐}	&	\text{microwave oven}	\\
			\hline
		\end{array}
	\end{align*}
while unrelated characters and words include
	\begin{align*}
		&\begin{array}{| p{2cm} | p{2cm} | p{3.5cm}  |}
			\hline
			\code{Char}	&	\text{香}	&	\text{fragrant}	\\
			\hline
			\code{Char}	&	\text{港}	&	\text{harbor}	\\
			\hline
			\code{Word}	&	\text{香港}	&	\text{Hong Kong}	\\
			\hline
		\end{array}
	\end{align*}
and
	\begin{align*}
		&\begin{array}{| p{2cm} | p{2cm} | p{3.5cm}  |}
			\hline
			\code{Char}	&	\text{幽}	&	\text{dark or quiet}	\\
			\hline
			\code{Char}	&	\text{默}	&	\text{silently or secretly}	\\
			\hline
			\code{Word}	&	\text{幽默}	&	\text{humor(ous)}	\\
			\hline
		\end{array}
	\end{align*}
In addition, the same sentence can sometimes be segmented differently and still remain grammatically 
correct, making CWS dependent on context. For example:
	\begin{align*}
		\begin{array}{| l | l |}
			\hline
			\text{結婚/的/和尚/没/結婚/的/學生。} & \text{A married monk and an unmarried 
												student.}	
			\\[2.5pt]
			\hline
			\text{結婚/的/和/尚没/結婚/的/學生。} & \text{Married and unmarried students.} 
			\\[2.5pt]
			\hline
		\end{array}
	\end{align*}
Additionally, there exists different labelling standards - all equally logical - to the same sentence. For 
example:
	\begin{align*}
		\begin{array}{| p{6cm} | p{6cm} |}
			\hline
			\text{案件/在/香港/區域/法院/審理。} & \text{The case is being heard in the }	\\[2.5pt]
			\hhline{|-|}
			\text{案件/在/香港/區域法院/審理。} & \text{District Court of Hong Kong SAR.}	\\[2.5pt]
			\hhline{|-|}
			\text{案件/在/香港區域法院/審理。} &	\\[2.5pt]
			\hline
		\end{array}
	\end{align*}
This makes the creation of a large scale CWS training corpus for deep learning-based segmentation
engine difficult.

Lastly, domain specific vocabulary imposes extra scalability challenges, for example
	\begin{align*}
		\begin{array}{| p{3.5cm} | p{3.5cm} |}
			\hline
			\text{量子引力}	&	\text{Quantum Gravity}
			\\
			\hline
			\text{林戴勝科}	&	\text{Wood Hoopoe}
			\\
			\hline
		\end{array}
	\end{align*}
A compromising solution is to train for high dimensional character embeddings in hopes of capturing 
the complex relationships between Chinese words and characters; however, downstream Euclidean 
models trained on Euclidean character embeddings have not been shown to match the performances 
of Euclidean models trained on word embeddings \cite{Chen2015b}.


\section{Hyperbolic Geometry}


\subsection{Riemannian Geometry}

First we briefly describe Riemannian geometry concepts crucial for understanding our algorithms. 
Interested readers can refer to \cite{Petersen2006} for a more in-depth view of the topic. An 
$n$-dimensional manifold $\mathcal{M}$ is a smooth space that is locally flat, so that at each point 
$\mathbf{x} \in \mathcal{M}$, we associate a tangent space $T_\mathbf{x}\mathcal{M}$. A manifold is 
Riemannian if it admits an inner product at each point (also known as a metric) 
$g_\mathbf{x}: T_\mathbf{x}\mathcal{M} \to T_\mathbf{x}\mathcal{M}$. Due to the smoothness of the 
manifold, one can admit smooth parametrized curves on them $\gamma : \mathbb{R} \to \mathcal{M}$. 
For simplicity we assume the domain of the parametrized curves is $[0,1]$\footnote{This is so that for a 
parametrized curve $\gamma$ on a manifold $\mathcal{M}$ that starts at $\mathbf{x}$ and ends at 
$\mathbf{y}$ we have $\gamma(0) = \mathbf{x}$ and $\gamma(1) = \mathbf{y}$}. The length of any 
parametrized curve $L$ is then determined by the metric: $L = \int_0^1 \sqrt{g_{\gamma(t)}(\gamma'(t), 
\gamma'(t))} \, dt$. The minimum-length parametrized curve connecting two points on the manifold 
$\mathbf{x}, \mathbf{y} \in \mathcal{M}$ is known as the geodesic between $\mathbf{x}$ and 
$\mathbf{y}$, and its length is known as the distance between points $\mathbf{x}$ and $\mathbf{y}$, 
denoted $d_\mathcal{M}(\mathbf{x}, \mathbf{y})$. 

It is often useful to transport vectors found in different tangent spaces in a parallel way to the same 
vector space so that we may operate on them using the same metric. This is known as parallel 
transport. Vectors in tangent spaces at any point can be parallel transported along geodesics to 
tangent spaces at other points such that they remain parallel (i.e. pointing in the same direction relative 
to the geodesic). This is done by breaking down the target vector into a linear combination of two 
orthogonal components: one pointing in the direction of the geodesic, and one pointing in the direction 
orthogonal to the geodesic. Parallel transport allows for more straight-forward addition of vectors found 
in different parts of the manifold. For a vector $\mathbf{v} \in T_\mathbf{x}\mathcal{M}$, we denote its 
parallel transport to the tangent space at another point $\mathbf{y} \in \mathcal{M}$ as $P_{\mathbf{x} 
\to \mathbf{y}}(\mathbf{v})$. 

Often one needs a convenient way to map from tangent spaces back to the manifold and vice versa. To 
this end, we incorporate the exponential and logarithmic mappings, which are functions 
$\exp_\mathbf{x} : T_\mathbf{x}\mathcal{M} \to \mathcal{M}$ and 
$\log_\mathbf{y}: \mathcal{M} \to T_\mathbf{y}\mathcal{M}$. Essentially, for any tangent vector 
$\mathbf{v} \in T_\mathbf{x}\mathcal{M}$, we can map $\mathbf{v}$ to a point that is $\|\mathbf{v}\|$ 
away from $\mathbf{x}$ in the direction of $\mathbf{v}$ by using the exponential mapping. The 
logarithmic mapping is the local inverse of the exponential mapping. 

As discussed in \cite{Ganea2018}, the exponential and logarithmic mappings can be used to define 
hyperbolic versions of transformations in models of hyperbolic geometry whose tangent space at the 
origin resembles $\mathbb{R}^n$. If  $f: \mathbb{R}^n \to \mathbb{R}^m$ is a Euclidean
transformation, we can define a hyperbolic version of $f$, denoted $f^{\otimes} : \mathcal{M}^n \to 
\mathcal{M}^m$ that maps from an $n$-dimensional manifold $\mathcal{M}^n$ to its corresponding 
$m$-dimensional manifold $\mathcal{M}^m$ by
	\begin{align*}
		f^{\otimes}(\mathbf{x}) &= \exp_\mathbf{0}(f(\log_\mathbf{0}(\mathbf{x})))
	\end{align*}
where $\exp_\mathbf{0}: T_{\mathbf{0}_m}\mathcal{M}^m \to \mathcal{M}^m$ and 
$\log_\mathbf{0}: \mathcal{M}^n \to T_{\mathbf{0}_n}\mathcal{M}^n$. 
	
					         
\subsection{Analytic Hyperbolic Geometry}

We want to establish an algebraic formalism for hyperbolic geometry that will help us perform 
operations in the same way that the vector-space structure of Euclidean geometry provides an 
algebraic formalism that allows the use of simple operations such as vector addition and scalar 
multiplication. 

As detailed in \cite{Ungar2005}, we can establish a non-associative\footnote{By non-associative, we 
mean that $a+(b+c) \neq (a+b)+c$ in general.} algebraic formalism that allows for operations 
analogous to vector addition and scalar multiplication. This formalism is known as the theory of 
gyrovector spaces, and uses concepts from analytic hyperbolic geometry\footnote{In analogy with 
Euclidean geometry, where analytic geometry provides an algebraic way to describe motions in 
(Euclidean) vector spaces, analytic hyperbolic geometry provides an algebraic way to describe 
motions in (hyperbolic) gyrovector spaces.}.

Essentially, a gyrovector space is like a vector space in that it is closed under its operations of scalar 
multiplication and gyrovector addition, and contains an identity element $\mathbf{0}$ and inverse 
elements $\ominus \mathbf{x}$\footnote{For a hyperbolic manifold $\mathcal{M}$, the inverse element 
of $\mathbf{x} \in \mathcal{M}$, denoted $\ominus \mathbf{x}$, is the point such that 
$\ominus \mathbf{x} \oplus \mathbf{x} = \mathbf{x} \oplus (\ominus \mathbf{x}) = \mathbf{0}$.}. Unlike 
vector spaces, gyrovector addition, denoted $\oplus$, is not associative, but it is 
\emph{gyroassociative} i.e. left-associative under the action of an automorphism known as a gyration. 
For any three gyrovectors $\mathbf{a}, \mathbf{b}, \mathbf{c}$, we have
	\begin{align*}
		\mathbf{a} \oplus (\mathbf{b} \oplus \mathbf{c}) &= (\mathbf{a} \oplus \mathbf{b}) \oplus
			\gyr[\mathbf{a}, \mathbf{b}]\mathbf{c}
	\end{align*}
where  $\gyr[\mathbf{a}, \mathbf{b}] \mathbf{c} = \ominus(\mathbf{a} \oplus \mathbf{b}) 
\oplus (\mathbf{a} \oplus (\mathbf{b} \oplus \mathbf{c}))$.

					         
\subsection{Models of Hyperbolic Geometry}

In this paper we will concern ourselves chiefly with two $n$-dimensional models of hyperbolic 
geometry: the Poincar\'e ball, denoted $\mathbb{D}^n$ and the hyperboloid model, denoted 
$\mathbb{H}^n$. A detailed discussion of models of hyperbolic geometry and their relationships to one 
another can be found in \cite{Cannon1997}.

\subsubsection{The Poincar\'e Ball Model of Hyperbolic Geometry}

The Poincar\'e ball of radius $c$ is a model of hyperbolic geometry defined by
	\begin{align*}
		\mathbb{D}^n_c &= \{ \mathbf{x} \in \mathbb{R}^n : \|\mathbf{x}\| < c\}.
	\end{align*}
its metric, denoted $g^{\mathbb{D}_c}$, is conformal to the Euclidean metric, denoted $g^E$, with a 
conformal factor $\lambda_\mathbf{x}^c$, i.e. 
$g_\mathbf{x}^{\mathbb{D}_c} = (\lambda_\mathbf{x}^c)^2g^E$, where
	\begin{align*}
		\lambda_\mathbf{x}^c = \frac{2}{1 - \frac{\|\mathbf{x}\|^2}{c^2}}
	\end{align*}
The Poincar\'e ball model of hyperbolic space forms a gyrovector space 
$(\mathbb{D}^n_c, \oplus_c, \otimes_c)$ with gyrovector addition $\oplus$ and scalar multiplication 
$\otimes$ given by \emph{M\"obius operations}.
As shown in \cite{Ungar2005}, for $\mathbf{x}, \mathbf{y} \in \mathbb{D}^n_c$, 
$r \in \mathbb{R}$, M\"obius addition and scalar multiplication are given by
	\begin{align*}
		\mathbf{x} \oplus_c \mathbf{y} &= 
			\frac{(1 + \frac{2}{c^2}\langle \mathbf{x}, \mathbf{y} \rangle + 
			\frac{1}{c^2}\|\mathbf{y}\|^2)\mathbf{x} +(1 - \frac{1}{c^2}\|\mathbf{x}\|^2)\mathbf{y}}{1 
			+  \frac{2}{c^2}\langle \mathbf{x}, \mathbf{y} \rangle + 
				\frac{1}{c^4} \|\mathbf{x}\|^2 \|\mathbf{y}\|^2}
		\\
		r \otimes_c \mathbf{x} &= c \tanh\left(r \tanh^{-1}\frac{\|\mathbf{x}\|}{c}\right)
			\frac{\mathbf{x}}{\|\mathbf{x}\|}
	\end{align*}
And its distance function is given by
	\begin{align*}
		d_{\mathbb{D}_c}(\mathbf{x}, \mathbf{y}) = 2c\tanh^{-1} 
			\left( \frac{1}{c} \| \ominus_c \mathbf{x} \oplus_c \mathbf{y}\| \right)
	\end{align*}
The exponential and logarithmic mappings for the Poincar\'e ball are derived in 
\cite{Ganea2018}\footnote{In \cite{Ganea2018}, they use the convention that 
$\frac{1}{\sqrt{c}}$ instead of $c$. We stick with the notation of \cite{Ungar2005} and use $c$, although 
in practice when implementing the algorithms, it is simpler to use $\frac{1}{\sqrt{c}}$.} and shown to be
	\begin{align*}
		\exp_\mathbf{x}^c &= \mathbf{x} \oplus_c \left( \tanh \left( 
			\frac{\lambda_\mathbf{x}^c\|\mathbf{v}\|}{2c}\right)
			\frac{c\mathbf{v}}{\|\mathbf{v}\|}\right)
		\\
		\log_\mathbf{x}^c(y) &= \frac{2c}{\lambda_\mathbf{x}^c} \tanh^{-1}\left( 
			\frac{1}{c}\|\ominus_c \mathbf{x} \oplus_c \mathbf{y}\|\right)
			\frac{\ominus_c \mathbf{x} \oplus_c \mathbf{y}}
			{\|\ominus_c \mathbf{x} \oplus_c \mathbf{y}\|}
	\end{align*}
\cite{Ganea2018} also shows that scalar multiplication can be defined using these two mappings as
$r \otimes_c \mathbf{x} = \exp_\mathbf{0}^c ( r \log_\mathbf{0}^c(\mathbf{x}))$, and parallel transport of 
a tangent space vector at the origin $\mathbf{v} \in T_\mathbf{0}\mathbb{D}^n_c$ to any other tangent 
space $T_\mathbf{x}\mathbb{D}^n_c$ becomes
	\begin{align*}
		P_{\mathbf{0} \to \mathbf{x}}^c(\mathbf{v}) &= \frac{\lambda_\mathbf{0}^c}
			{\lambda_\mathbf{x}^c}\mathbf{v}
	\end{align*}
Exponential and logarithmic mappings can also be used to define M\"obius matrix-vector multiplication 
and bias translations:
	\begin{align*}
		M^{\otimes_c}(\mathbf{x}) &= 
			\begin{cases}
				c \tanh\left( \frac{\|M\mathbf{x}\|}{\|\mathbf{x}\|} 
				\tanh^{-1} \left( \frac{\|\mathbf{x}\|}{c}  \right)\right) 
				\frac{M\mathbf{x}}{\|M\mathbf{x}\|} &\text{ if } M\mathbf{x} \neq \mathbf{0}
				\\
				\mathbf{0} &\text{ if } M\mathbf{x} = \mathbf{0}
			\end{cases}
		\\
		\mathbf{x} \oplus_c \mathbf{b} &= \exp^c_{\mathbf{x}} ( 
			P^c_{\mathbf{0} \to \mathbf{x}}(\log^c_\mathbf{0}(\mathbf{b}))).
	\end{align*}	
Finally, we note here that many of the formulae above become greatly simplified by setting the radius 
of the ball $c = 1$, so we adopt this convention unless otherwise stated. We will, however, continue to 
use $\oplus_c$ to denote M\"obius addition and $\otimes_c$ to denote M\"obius scalar multiplication 
in order to draw attention to the fact that these operations are performed in the Poincar\'e ball model 
of hyperbolic geometry. 

					         
\subsubsection{The Hyperboloid Model of Hyperbolic Geometry}

A detailed discussion on the hyperboloid model can be found in \cite{Reynolds1993}. The hyperboloid 
model is an $n$-dimensional manifold embedded in $(n+1)$-dimensional Minkowski space, denoted 
$\mathbb{R}^{(n,1)}$ which is the usual Euclidean $(n+1)$-dimensional vector space endowed with the 
Lorentzian inner product, which, for $\mathbf{x}, \mathbf{y} \in \mathbb{R}^{(n,1)}$ is given 
by\footnote{Note that in much of the literature, this Lorentzian inner product is written as $\langle 
\mathbf{x}, \mathbf{y} \rangle_\mathcal{L} = -x_1y_1 + \sum x_iy_i$. This is a matter of notation; the 
results presented here are valid either way, but it is important to keep track of which kind of Lorentzian 
inner product is being used when implementing these algorithms.}
	\begin{align*}
		\displaystyle \langle \mathbf{x}, \mathbf{y} \rangle_\mathcal{L} &= \sum_{i=1}^n x_iy_i -
			x_{n+1}y_{n+1}
	\end{align*} 
and the Hyperboloid model is given by
	\begin{align*}
		\mathbb{H}^n &= \{ \mathbf{x} \in \mathbb{R}^{(n,1)} :  
			\langle \mathbf{x}, \mathbf{x} \rangle_\mathcal{L} = -1, x_{n+1} > 0\}
	\end{align*}
and its metric is also given by the Lorentzian inner product: 
$g^\mathbb{H}_\mathbf{x}(\mathbf{u}, \mathbf{v}) = \langle \mathbf{u}, \mathbf{v} \rangle_\mathcal{L}$. 
The distance between two points is given by
	\begin{align*}
		d_\mathbb{H}(\mathbf{u}, \mathbf{v}) &= \cosh^{-1}(
			-\langle \mathbf{u}, \mathbf{v} \rangle_\mathcal{L})
	\end{align*}	
If two points on the hyperboloid $\mathbf{x}, \mathbf{y} \in \mathbb{H}^n$ are connected by a geodesic 
that points in the direction $\mathbf{v}$ of length $\|\mathbf{v}\|$, then we can consider a unit vector in 
the same direction $\hat{\mathbf{v}}$ and define the parallel transport of tangent space vectors 
$\mathbf{w} \in T_\mathbf{x}\mathbb{H}^n$ to the tangent space 
$T_\mathbf{y}\mathbb{H}^n$.	
	\begin{align*}
		P_{\mathbf{x} \to \mathbf{y}}(\mathbf{w}) &= \langle \mathbf{w}, 
			\mathbf{\hat v}\rangle_\mathcal{L} (\sinh(\|\mathbf{v}\|) \mathbf{x} + 
			\cosh(\|\mathbf{v}\|)\mathbf{\hat v}) + (\mathbf{w} - \langle \mathbf{w}, 
			\mathbf{\hat v} \rangle_\mathcal{L} \mathbf{\hat v})
	\end{align*}
Finally, we will often need to project vectors in the ambient Minkowski space onto the tangent spaces 
of the hyperboloid. To do this, suppose we have a point on the hyperboloid 
$\mathbf{x} \in \mathbb{H}^n$, and a vector in the ambient Minkowski space 
$\mathbf{v} \in \mathbb{R}^{(n,1)}$. We can project $\mathbf{v}$ onto the tangent space 
$T_\mathbf{x}\mathbb{H}^n$ using the following:
	\begin{align*}
		\proj_\mathbf{x}(\mathbf{v}) &= \mathbf{v} + \langle \mathbf{x}, \mathbf{v} 
			\rangle_\mathcal{L} \mathbf{x}.
	\end{align*}


\section{Hyperbolic Neural Network Structures}


\subsection{Hyperboloid Char2Vec}

The Euclidean skip-gram architecture found in \cite{Mikolov2013} can be summarized as follows. 
Suppose we have a dictionary of words $\mathcal{V}$. Given a continuous stream of text 
$T = (w_0, w_1, ..., w_n)$, where $w_i \in \mathcal{V}$, skip-gram learns a vector representation in 
Euclidean space for each word by using it to predict surrounding words. Given a center word $w_k$, 
and $2\mu$ surrounding words (known as a context) given by $C = (w_{k-\mu},...,w_{k-1}, 
w_{k+1}, ..., w_{k+\mu})$, the task is to predict $C$ from $w_k$. In order to train this model, we 
incorporate the negative sampling proposed in \cite{Mikolov2013b}. The center word and the context 
words are parametrized as two layers of a neural network, where the first layer represents the 
projection of the center word, and the output layer represents the context words. Suppose we wish to 
form embeddings of dimension $d$. Then we can parametrize these layers using matrices $A \in 
\mathbb{R}^{d \times |\mathcal{V}|}$ for the first layer and $B \in \mathbb{R}^{d \times |\mathcal{V}|}$. 
We index the columns of each matrix using the words from the dictionary $w_i \in \mathcal{V}$, i.e. 
$A_{w_i}$ and $B_{w_i}$. Suppose our center word is $w_k$. Let $w_{k+j}$ be some context word. 
Then negative sampling chooses $m$ random noise samples $\{w^1, ..., w^m\}$, and seeks to 
minimize the loss function given by
	\begin{align*}
		\displaystyle L_{w_k,w_{k+j}} (A, B) &=  \prod_{i=0}^m p(y_i | w^i, w_k)  
		\\
		&\propto \prod_{i=0}^m \sigma((-1)^{1-y_i}  \langle A_{w_k}, B_{w_{k+j}} \rangle)
	\end{align*}
where labels are given by $y_0 =1$ and $y_i = 0$ otherwise. 

We wish to create word embeddings that capture both symmetric and asymmetrical relationships, and 
that efficiently model hierarchical relationships between words and characters. For this we turn to 
hyperbolic geometry. Since computing gradients on the hyperboloid model is easier than the Poincar\'e 
ball, \cite{Wilson2018}\cite{Nickel2018} we follow \cite{Leimeister2018} and use the hyperboloid model. 
To create a hyperboloid version of this loss function, \cite{Leimeister2018} proposes replacing the 
above Euclidean dot product with the Lorentzian product with an additive shift\footnote{The additive 
shift is placed because the hyperboloid restricts the Minkowski space so that for all $\mathbf{x},
\mathbf{y} \in \mathbb{H}^n$, we have that $\langle \mathbf{x}, \mathbf{y} \rangle_\mathcal{L} \geq -1$ 
with equality iff. $\mathbf{x} = \mathbf{y}$ -- the additive shift ensures we don't end up with negative 
probabilities. }:
	\begin{align*}
		\displaystyle L_{w_k,w_{k+j}} (A, B) &\propto \prod_{i=0}^m \sigma((-1)^{1-y_i}(\langle 
			A_{w_k}, B_{w_{k+j}} \rangle_\mathcal{L} + \theta))
	\end{align*}
We then optimize this loss function using Riemannian Stochastic Gradient Descent 
\cite{Bonnabel2011}. The details are given in the appendix. 

					         
\subsection{Hyperbolic Transformer}

Hyperboloid word embeddings are used downstream in a hyperbolic intent classification model. We 
use the Transformer architecture proposed in \cite{Vaswani2017}. 

The core of the Euclidean version of Transformer is the operation called Scaled Dot Product attention. 
An input sequence is converted into three vectors, $Q,K \in \mathbb{R}^{d_k}$, and $V \in 
\mathbb{R}^{d_v}$\footnote{Usually $d_k = d_v$} and the attention mechanism is then computed as 
follows: 
	\begin{align*}
		\text{Attention}(Q, K, V) = \text{softmax}\left(\frac{QK^T}{\sqrt{d_k}}\right)V
			\label{attention}
	\end{align*}
	
The 3 vectors are split into $h$ heads via linear projections $W^Q_i, W^K_i$ and $W^V_i$  with their 
results concatenated at the end, and then projected again by an output projection $W^O$. This allows 
the model to attend to multiple information subspaces simultaneously, resulting in better generalization. 
This is called Multi-head Attention, and is computed as follows: for $i = 1,...,h$,
	\begin{align*}
		\text{head}_i &= \text{Attention}(QW^Q_i, KW^K_i, VW^V_i)
		\\		
		\text{Multi-head} &= \text{concat}\left[\text{head}_1, \hdots ,
			\text{head}_h\right]W^O
	\end{align*}
The output of Multihead Attention is then fed into a fully connected layer. We stack $N$ layers of 
Transformer. To perform intent-classification, the final output of the Transformer is max pooled, fed to 
another fully connected feed-forward layer, and then finally softmaxed. 

The Transformer itself does not capture information about the position of elements in the input 
sequence. In order to encode the positioning of these elements, we add a positional encoding to the 
input sequence of dimension $d$:
	\begin{align*}
		&PE_{pos,2i} = \sin\left( \frac{pos}{10000^{\frac{2i}{d}}}\right),
		&PE_{pos,2i+1} = \cos\left( \frac{pos}{10000^{\frac{2i}{d}}}\right).
	\end{align*}

The hyperbolic version of the Transformer architecture replaces Euclidean inputs with hyperbolic ones. 
Due to the simplicity of the expression for parallel transport of vectors in the Poincar\'e ball, we adopt 
this model of hyperbolic geometry. This means that our hyperboloid word-embeddings need to be 
transformed from the hyperboloid to the Poincar\'e model. This transformation is detailed in appendix A.

The positional encodings live in $\mathbb{R}^n$. Since this is equivalent to saying that they live in the 
tangent space at the origin of the Poincar\'e ball, we use the exponential mapping to map the 
encodings from this tangent space back down to the Poincar\'e ball, and then gyroadd the inputs and 
the positional encodings
	\begin{align*}
		\code{transformer\_input} = \code{input} \oplus_c \exp_\mathbf{0}^c(PE).
	\end{align*}

\cite{Gulcehre2018} suggests using a softmaxed hyperbolic distance function scaled by a temperature 
and translated by a bias, followed by a hyperbolic midpoint function, such as the Einstein midpoint 
function (see \cite{Ungar2005} ch. 6.20.1 Thm 6.87) to replace scaled dot product attention. We find, 
however, that leveraging parallel transport is a more numerically stable and successful paradigm to 
adopt in this case. We can parallel transport the query, key and value vectors to the origin using the 
logarithmic and exponential mappings, and then proceed to compute the scaled dot product attention 
using the standard Euclidean operations. This is viable because the tangent space at the origin of the 
Poincar\'e ball resembles $\mathbb{R}^n$. 

To split our $n$-dimensional hyperbolic queries, keys and values into $m$ heads, each with 
dimension $h$ we need to use a single hyperbolic matrix multiplication first and then split into heads. 
For the input $\mathbf{x} \in \mathbb{D}^n_c$, we use a matrix $W \in \mathbb{R}^{3h \times n}$ to 
create a concatenated vector of heads, which we can then split:
	\begin{align*}
		W \otimes_c \mathbf{x} &= \code{heads} = [\code{head}_1, ..., \code{head}_m]
	\end{align*}
These $\code{head}_i$ are then fed into the multihead attention. The result of each head must then be 
concatenated. However, straightforward concatenation is not a valid operation in hyperbolic space. 
Consequently, we must instead use hyperbolic matrix multiplication on each head individually and then 
gyroadd the results together at the end. So for $i = 1,...,m$, we have
	\begin{align*}
		M_i \otimes_c \code{head}_i &= \code{result}_i
		\\
		\code{result}_1 \oplus_c ... \oplus_c \code{result}_m &= \code{concatenated\_result}.
	\end{align*}	
To modify the feed-forward neural network, we simply change it to a hyperbolic feed-forward neural 
network with two layers. For a hyperbolic neural network input $\mathbf{x}$, we have
	\begin{align*}
		M_1 \otimes_c \mathbf{x} \oplus_c \mathbf{b}_1 &= \mathbf{h}
		\\
		M_2 \otimes_c \mathbf{h} \oplus_c \mathbf{b}_2 &= \hat{\mathbf{y}}
	\end{align*}
The output of the hyperbolic neural network is then max pooled and passed through a hyperbolic 
logistic regression algorithm in order to classify intents. Hyperbolic logistic regression is derived in 
\cite{Ganea2018} and detailed in the appendix.


\section{Experiments}
We have collected 16332 user utterances in the music domain. Each user utterance is a text-based 
voice command for an Alexa-like machine, with a labelled user intent.

Many of these utterances carry multiple commands such as \code{query\_song + increase\_volume,} 
thus inducing hierarchical relationships between composite intents and singular intents, warranting the 
use of hyperbolic machine learning. There are 125 intents in total. The list of intents can be found in the 
appendix section D.

We held out 15\% of the dataset for evaluating intent classification accuracy.

Both Euclidean and hyperbolic character skip-gram embeddings are trained on a Chinese corpus of 
newspapers from Linguistic Data Consortium (LDC) of roughly 368 million characters \cite{LDC}.

For Euclidean Transformers, we used RMSProp optimisers \cite{HintonRMSProp} with a learning rate 
of 0.0001; hyperbolic Transformers were mostly optimised with RMSProp at a learning rate of 0.001, 
since most of the kernel matrices are Euclidean variables. The hyperbolic biases were optimised with 
Riemannian SGD \cite{Bonnabel2011} at a learning rate of 0.05. In all cases, Transformers consist of 3 
multihead-attention layers, each of which splits its inputs into a total of 16 heads. 

We trained 2 character-based Euclidean models, one 128-dimensional and the other 256-dimensional, 
while both character-based hyperbolic models are 100-dimensional. Both Euclidean models use a 
dropout of 20\%. One of the hyperbolic models uses no dropout, and the other one uses a dropout of 
30\%. 

To validate hyperbolic geometry's suitability for modelling Chinese word-character relationships, we 
created another 256-dimensional Euclidean Transformer trained with word tokens instead of character 
tokens. The words were segmented with bi-directional LSTM + CRF \cite{Chen2015} and a heavily 
music-domain-optimised post-processing dictionary. The word segmentation engine was trained on 
People's Daily corpus \cite{Yu2001}, and the word-embeddings were also trained using Euclidean 
Skip-gram on the LDC data set. 

All models were restarted once \cite{Loshchilov2017}; we observe an accuracy boost from restarts for 
all models.

					         
\begin{table}
	\begin{tabular}{|l||c||c|}
		\hline
			&	Intent Classification Accuracy	&	Cross-Entropy Loss\\[2pt]
		\hline
		\code{Eucl\_TRF c2v 128D w/ 20\% dropout} &	94.0\%	&	0.4808	\\[2pt]
		\hline
		\code{Eucl\_TRF c2v 256D w/ 20\% dropout}	&	94.8\%	&	0.4330	\\[2pt]
		\hline
		\code{Eucl\_TRF w2v 256D w/ 20\% dropout}	&	95.6\%	&	0.1684	\\[2pt]
		\hhline{|=||=||=|}
		\code{Hyp\_TRF c2v 100D no dropout}	&	96.2\%	&	0.1729	\\[2pt]
		\hline
		\code{Hyp\_TRF c2v 100D w/ 30\% dropout}	&	\textbf{96.9\%}	& \textbf{0.1226}\\[2pt]
		\hline
	\end{tabular}
	\caption{Intent Classification Accuracy and Cross-Entropy Loss for five different versions of the 
	Transformer Model for intent classification. Despite being a lower-dimensional model, the 
	hyperbolic Transformer outperforms its Euclidean counterpart.}
	\label{results_table}
\end{table}

\section{Results}

The results are shown in table \ref{results_table}. We first note that, as expected, the word-based 
Euclidean Transformer outperforms the character-based Euclidean Transformers. This is because 
training with word-based tokens allows the algorithm to model more complex and nuanced 
relationships between characters by simply identifying them as belonging to certain word tokens, which 
helps downstream to improve the accuracy of intent classification. 

Of particular interest to us, however, is the marked improvement in accuracies of 
hyperbolic Transformers over their Euclidean counterparts. We believe that since words and characters 
form an obvious hierarchical relationship, and that words often have asymmetrical relationships between 
each other, that hyperbolic geometry is naturally better suited to encoding characters in a natural 
language understanding task, thus explaining these improved results. 

Additionally, the fact that character-based hyperbolic intent classification (i.e. that makes use of 
character-based embeddings) still outperforms word-based Euclidean intent classification is a 
promising sign. It indicates that leveraging the power of hyperbolic representations of natural language 
can capture the hierarchical and asymmetrical relationships between words and characters well 
enough to circumvent the need for CWS altogether. This indeed is a promising research direction.


\section{Conclusion}

To the best of our knowledge, we have been the first to use hyperbolic embeddings in a 
downstream hyperbolic deep-learning task. Our results show that hyperbolic character-based 
intent-classification outperforms its character-based Euclidean counterpart, giving us confidence that 
hyperbolic embeddings and hyperbolic deep learning captures hierarchical and asymmetrical 
relationships in the Chinese language better than Euclidean embeddings and deep learning do. 
Additionally, we found that hyperbolic character-based intent-classification even outperforms Euclidean 
word-based intent-classification, which itself requires the use of state-of-the-art CWS. As we have 
discussed, CWS is a difficult and as-yet unscalable task, yet our results indicate that hyperbolic 
character-based deep-learning may be able to dispense with the need for this difficult task altogether. 
In our opinion, this indicates that hyperbolic deep learning merits further research, particularly in the 
realm of Chinese NLU. 



\section*{Appendix}


\subsection*{A. Relationship between Poincar\'e ball and Hyperboloid Model}

\begin{figure}[ht!]
	\centering
	\includegraphics[width=105mm]{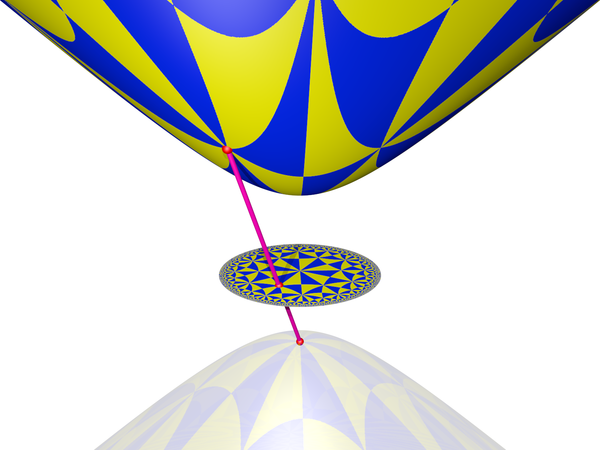}
	\caption{A hyperboloid model of hyperbolic space can be arrived at by placing a Poincar\'e disk 
	at the plane of symmetry between a two sheet hyperboloid, and then projecting points from the 
	Poincar\'e disk onto the upper sheet of a two sheet hyperboloid using the vertex of the lower 
	sheet as the center of projection.  This image is taken from \cite{vonGagern2014}.}
	\label{hyp_model_poincare_ball}
	\end{figure}

The hyperboloid model can be visualized as the top part of a 2-sheet hyperboloid living in 
$\mathbb{R}^3$. Its plane of symmetry is the $xy$ plane. We can place a Poincar\'e disk of radius $c$ 
(usually $c = 1$) on this plane of symmetry centered at the origin. We can map points from the 
Poincar\'e disk to the hyperboloid by projecting them from the disk to the hyperboloid with the vertex 
of the bottom part of the 2-sheet hyperboloid acting as the center of projection. This is shown in figure 
\ref{hyp_model_poincare_ball}. 

This provides us with a simple way of determining a mapping and its inverse mapping from one model 
to the other. Suppose we wish to map from the hyerboloid model to the Poincar\'e disk. To do this, we 
simply define a projection function $\rho: \mathbb{H}^n \to \mathbb{D}^n$. Now suppose we choose a 
point $\mathbf{x}$ on the hyperboloid. Then $\mathbf{x} \in \mathbb{R}^{(n,1)}$ and $x_{n+1} > 0$, 
and we can then define our projection of this point from the hyperboloid to the Poincar\'e disk as 
follows:
	\begin{align*}
		\rho(\mathbf{x}) = \rho(x_1, x_2, ..., x_n, x_{n+1})
		&= \frac{1}{x_{n+1} + 1} (x_1, x_2, ..., x_n)
	\end{align*}

and its inverse is given for any $\mathbf{y} \in \mathbb{D}^n$:
	\begin{align*}
		\rho^{-1}(\mathbf{y}) &= \frac{2}{1 - \|\mathbf{y}\|} \left( y_1, y_2, ..., y_n, 
			\frac{1 + \|\mathbf{y}\|^2}{2} \right)
	\end{align*}

In this way we can transform from one model to the other and back again at our convenience. This is 
useful as it allows us to leverage the advantages of both models where needed. 


\subsection*{B. Riemannian Stochastic Gradient Descent of the Hyperboloid Model}

This process is detailed in \cite{Leimeister2018}. Gradients in Minkowski space $\mathbb{R}^{(n,1)}$ 
are taken with regard to the characteristics of the Lorentzian inner product. For a differentiable function 
$f: \mathbb{R}^{(n,1)} \to \mathbb{R}$, the gradient is given by
	\begin{align*}
		\nabla f = \left( \frac{\partial f}{\partial x_0}, ..., \frac{\partial f}{\partial x_{n-1}}, 
		- \frac{\partial f}{\partial x_n} \right)
	\end{align*}
These gradients can be projected into the tangent space $T_\mathbf{x}\mathbb{H}^n$ at a parameter 
point $\mathbf{x}$ to form Riemannian gradients. For the first layer parametrized by $A_c$, we have 
the following gradient equation for the log-likelihood
	\begin{align*}
		\displaystyle \nabla_{A_c}^{\mathbb{R}^{(n,1)}} \log L_{c, w_0}(A, B) &= \sum_{i=0}^k
			(y_i - \sigma(\langle A_c, B_{w_i} \rangle_\mathcal{L} + \theta)) B_{w_i}
	\end{align*}
For the second layer, $B_w$, first consider the set consisting of the positive sample $w_0$ and all the 
negative samples $w_1,...,w_k$ given by
	\begin{align*}
		\mathcal{S}_c &= \{w_0, w_1, ..., w_k \}.
	\end{align*}
Given any word $w \in \mathcal{V}$, let $N_{w,\mathcal{S}_c}$ be the number of times $w$ appears 
in $\mathcal{S}_c$, and let
	\begin{align*}
		y(w) = 
			\begin{cases}
				1 &\text{ if } w = w_0
				\\
				0 &\text{ if } w \in \{w_1, ..., w_k\}
			\end{cases}
	\end{align*}
Then the gradient is given by
	\begin{align*}
		\nabla_{B_w}^{\mathbb{R}^{(n,1)}}  \log L_{c, w_0}(A, B) &= N_{w,\mathcal{S}_c}
			(y(w) - \sigma(\langle A_c, B_{w_i} \rangle_\mathcal{L} + \theta)) A_c
	\end{align*}
Next we project these gradients onto the tangent spaces of the hyperboloid. Recall that for a point 
$\mathbf{x}$ on the hyperboloid and a vector $\mathbf{v}$ in the ambient Minkowski space, we can 
project $\mathbf{v}$ onto the tangent space at $\mathbf{x}$ using	
	\begin{align*}
		\proj_\mathbf{x}(\mathbf{v}) &= \mathbf{v} + \langle \mathbf{x}, \mathbf{v} 
			\rangle_\mathcal{L} \mathbf{x}
	\end{align*}
Then our projected gradients become
	\begin{align*}
		\nabla_{A_c}^{\mathbb{H}^n} \log L_{c, w_0}(A, B)&= \proj_{A_c}\left(\log L_{c, w_0}(A, B)
			\right)
		\\
		\nabla_{B_w}^{\mathbb{H}^n} \log L_{c, w_0}(A, B)&= \proj_{B_w}\left(\log L_{c, w_0}(A, B)
			\right)
	\end{align*}
We then optimize the projected gradients using an exponential mapping on the negative gradient 
scaled by a learning factor $\eta$:
	\begin{align*}
		A_c^\text{new} &= \exp_{A_c^\text{old}}(-\eta \nabla_{A_c^\text{old}}^{\mathbb{H}^n} 
			\log L_{c, w_0}(A, B))
		\\
		B_w^\text{new} &= \exp_{B_w^\text{old}}(-\eta \nabla_{B_w^\text{old}}^{\mathbb{H}^n} 
			\log L_{c, w_0}(A, B))
	\end{align*}
	
	
\subsection*{C. Hyperbolic Logistic Regression}

\cite{Ganea2018} describes a version of hyperbolic multiclass logistic regression (hMLR) that is 
applicable to the Poincar\'e ball. However, since the models of hyperbolic space are all isometric to 
each other, it is not difficult to determine how to perform hMLR on any other model of hyperbolic space. 

Suppose we have $K$ classes, $\{1, ..., K\}$. Euclidean MLR learns a margin hyperplane for each 
class using softmax probabilities i.e. for all $k \in \{1, ..., K\}, b_k \in \mathbb{R}$ and 
$\mathbf{x}, \mathbf{a} \in \mathbf{R}^n$, we have
	\begin{align*}
		p(y = k| \mathbf{x}) \propto \exp(\langle \mathbf{a}_k, \mathbf{x} \rangle - b_k )
	\end{align*}
	
This can be formulated from the perspective of measuring distances to marginal hyperplanes. A 
hyperplane can be defined by a normal nonzero vector $\mathbf{a} \in \mathbb{R}^n$ and a scalar shift 
$b \in \mathbb{R}$:
	\begin{align*}
		H_{\mathbf{a}, b} &= \{ \mathbf{x} \in \mathbb{R}^n : 
			\langle \mathbf{a}, \mathbf{x} \rangle - b = 0 \}
	\end{align*}
	
One can think of points in space $\mathbf{x} \in \mathbb{R}^n$ in relation to the hyperplane 
$H_{\mathbf{a}, b}$ by examining the points in relation to the orientation of the hyperplane (as oriented 
by the normal vector $\mathbf{a}$) and its distance to the hyperplane (scaled by the magnitude of the 
normal vector $\|\mathbf{a}\|)$:
	\begin{align*}
		\langle \mathbf{a}, \mathbf{x} \rangle - b = \text{sign}(\langle \mathbf{a}, \mathbf{x} \rangle 
			- b)\|\mathbf{a}\| d(\mathbf{x}, H_{\mathbf{a},b})
	\end{align*}
	
We can substitute this into the equation for MLR:
	\begin{align*}
		p(y = k| \mathbf{x}) &\propto \exp(\langle \mathbf{a}_k, \mathbf{x} \rangle - b_k )
		\\
		&= \exp( \text{sign}(\langle \mathbf{a}_k, \mathbf{x} \rangle 
			- b_k)\|\mathbf{a}_k\| d(\mathbf{x}, H_{\mathbf{a}_k,b_k}) )
	\end{align*}
	
We then reformulate by absorbing the bias term into the point, which creates a new definition for the hyperplane, where for $\mathbf{p} \in \mathbb{R}^n$
	\begin{align*}
		\tilde{H}_{\mathbf{a}, \mathbf{p}} &= \{ \mathbf{x} \in \mathbb{R}^n : 
			\langle -\mathbf{p} +\mathbf{x} , \mathbf{a} \rangle = 0 \}
	\end{align*}
	
where $\tilde{H}_{\mathbf{a}, \mathbf{p}} = H_{\mathbf{a}, \langle \mathbf{a}, \mathbf{p} \rangle}$. 
Then setting $b_k = \langle \mathbf{a}_k, \mathbf{p}_k \rangle$, we can rewrite MLR
	\begin{align*}
		p(y = k| \mathbf{x}) &\propto \exp( \text{sign}(\langle \mathbf{a}_k, \mathbf{x} \rangle 
			- b_k)\|\mathbf{a}_k\| d(\mathbf{x}, 
			H_{\mathbf{a}_k,\langle \mathbf{a}_k, \mathbf{p}_k \rangle}) )
		\\
		&= \exp( \text{sign}(\langle \mathbf{a}_k, \mathbf{x} \rangle 
			- \langle \mathbf{a}_k, \mathbf{p}_k \rangle)\|\mathbf{a}_k\| 
			d(\mathbf{x}, H_{\mathbf{a}_k,b_k}) )
		\\
		&=  \exp( \text{sign}(\langle -\mathbf{p}_k + \mathbf{x}, \mathbf{a}_k \rangle )
			\|\mathbf{a}_k\| d(\mathbf{x}, \tilde{H}_{\mathbf{a}_k,\mathbf{p}_k}) )
	\end{align*}
	
which we can then convert to a hyperbolic version in a manifold $\mathcal{M}$ with metric $g$:
	\begin{align*}
		\tilde{H}^\mathcal{M}_{\mathbf{a}, \mathbf{p}} &= \{ \mathbf{x} \in \mathcal{M} : 
			\langle \ominus \mathbf{p} \oplus \mathbf{x}, \mathbf{a} \rangle = 0 \}
		\\
		p(y = k| \mathbf{x}) &\propto \exp\left( \text{sign}(\langle \ominus \mathbf{p}_k \oplus 
		\mathbf{x}, \mathbf{a}_k \rangle )\sqrt{g_{\mathbf{p}_k}(\mathbf{a}_k, \mathbf{a}_k)} 
		d_\mathcal{M}(\mathbf{x}, \tilde{H}_{\mathbf{a}_k,\mathbf{p}_k}) \right)
	\end{align*}
In particular, for a Poincar\'e ball of dimension $n$ and radius $c$, denoted $\mathbb{D}^n_c$:
	\begin{align*}
		p(y = k| \mathbf{x}) &\propto \exp\left( c\lambda^c_{\mathbf{p}_k}\|\mathbf{a}_k\| 
			\sinh^{-1} \left( \frac{2 \langle \ominus \mathbf{p}_k \oplus \mathbf{x}, 
			\mathbf{a}_k \rangle}{c(1- \frac{1}{c^2} \|\ominus \mathbf{p}_k \oplus \mathbf{x} \|^2)
			\|\mathbf{a}_k\|} \right)\right)
	\end{align*}
and this can be optimized using Riemannian optimization. 


\subsection*{D. List of Intent Types}

\code{add\_playing\_song\_to\_blacklist}  \\
\code{add\_playing\_song\_to\_blacklist+query\_song\_hit+play\_song}  \\
\code{add\_preference}  \\
\code{add\_preference+play\_song}  \\
\code{add\_preference+query\_song\_hit+play\_song}  \\
\code{add\_to\_blacklist}  \\
\code{add\_to\_favourite}  \\
\code{add\_to\_favourite+play\_song}  \\
\code{add\_to\_favourite+sort\_playlist}  \\
\code{add\_to\_favourite+sort\_playlist+play\_song}  \\
\code{add\_to\_playing\_list}  \\
\code{add\_to\_playing\_list+play\_song}  \\
\code{add\_to\_playing\_list+sort\_playlist}  \\
\code{add\_to\_playing\_list\_next}  \\
\code{add\_to\_playlist}  \\
\code{add\_to\_playlist+play\_song}  \\
\code{add\_to\_playlist+sort\_playlist}  \\
\code{add\_to\_playlist+sort\_playlist+play\_song}  \\
\code{create\_playlist}  \\
\code{create\_playlist+play\_song}  \\
\code{create\_playlist\_from\_current}  \\
\code{destroy\_playlist}  \\
\code{fastbackward\_song}  \\
\code{fastforward\_song}  \\
\code{list\_favourite}  \\
\code{list\_playing\_list}  \\
\code{list\_playlist}  \\
\code{loop\_play\_album}  \\
\code{loop\_play\_playing\_list}  \\
\code{loop\_play\_playlist}  \\
\code{loop\_play\_song}  \\
\code{move\_from\_playlist}  \\
\code{move\_from\_playlist+play\_song}  \\
\code{music\_pause}  \\
\code{next\_song}  \\
\code{next\_song+loop\_play\_song}  \\
\code{others}  \\
\code{pause\_song}  \\
\code{pause\_song+play\_song}  \\
\code{play\_random}  \\
\code{play\_song}  \\
\code{play\_song\_get\_album}  \\
\code{play\_song\_get\_album+loop\_play\_album}  \\
\code{play\_song\_get\_album+loop\_play\_playing\_list}  \\
\code{play\_song\_get\_artist+add\_to\_blacklist}  \\
\code{play\_song\_get\_artist+loop\_play\_playing\_list}  \\
\code{play\_song\_get\_genre+add\_to\_blacklist}  \\
\code{play\_song\_get\_song}  \\
\code{play\_song\_get\_song+add\_to\_blacklist}  \\
\code{play\_song\_get\_song+add\_to\_playlist}  \\
\code{play\_song\_get\_song+query\_song}  \\
\code{previous\_song}  \\
\code{query\_favourite+play\_song}  \\
\code{query\_hit\_song+add\_to\_favourite}  \\
\code{query\_lyric}  \\
\code{query\_lyric+add\_to\_favourite}  \\
\code{query\_lyric+add\_to\_playing\_list}  \\
\code{query\_lyric+add\_to\_playlist}  \\
\code{query\_lyric+play\_song}  \\
\code{query\_lyric\_get\_mood}  \\
\code{query\_playlist}  \\
\code{query\_playlist+play\_song}  \\
\code{query\_preference}  \\
\code{query\_preference+add\_to\_favourite}  \\
\code{query\_preference+add\_to\_playing\_list}  \\
\code{query\_preference+add\_to\_playlist}  \\
\code{query\_preference+play\_song}  \\
\code{query\_song}  \\
\code{query\_song+add\_to\_blacklist}  \\
\code{query\_song+add\_to\_favourite}  \\
\code{query\_song+add\_to\_favourite+play\_song}  \\
\code{query\_song+add\_to\_playing\_list}  \\
\code{query\_song+add\_to\_playing\_list\_next+query\_song}  \\
\code{query\_song+add\_to\_playlist}  \\
\code{query\_song+add\_to\_playlist+loop\_play\_playlist}  \\
\code{query\_song+clock\_alarm\_set}  \\
\code{query\_song+loop\_play\_playing\_list}  \\
\code{query\_song+loop\_play\_song}  \\
\code{query\_song+play}  \\
\code{query\_song+play\_song}  \\
\code{query\_song+play\_song+add\_to\_playlist}  \\
\code{query\_song+play\_song+query\_song+play\_song+play\_song}  \\
\code{query\_song+vol\_max+play\_song}  \\
\code{query\_song\_get\_album}  \\
\code{query\_song\_get\_artist}  \\
\code{query\_song\_get\_genre}  \\
\code{query\_song\_hit}  \\
\code{query\_song\_hit+add\_to\_favourite}  \\
\code{query\_song\_hit+add\_to\_playing\_list}  \\
\code{query\_song\_hit+add\_to\_playlist}  \\
\code{query\_song\_hit+create\_playlist}  \\
\code{query\_song\_hit+play\_song}  \\
\code{query\_song\_hit+play\_song+stop\_song\_after}  \\
\code{query\_song\_hit+query\_song+play\_song}  \\
\code{query\_song\_latest}  \\
\code{query\_song\_latest+play\_song}  \\
\code{query\_song\_latest\_album+play\_song}  \\
\code{query\_song\_playlist}  \\
\code{query\_song\_preference+play\_song}  \\
\code{query\_song\_random}  \\
\code{query\_song\_random+add\_to\_favourite}  \\
\code{query\_song\_random+add\_to\_playing\_list}  \\
\code{query\_song\_random+add\_to\_playlist}  \\
\code{query\_song\_random+clock\_alarm\_set}  \\
\code{query\_song\_random+play\_song}  \\
\code{query\_song\_random+play\_song+add\_to\_favourite}  \\
\code{remove\_from\_blacklist}  \\
\code{remove\_from\_playing\_list}  \\
\code{remove\_from\_playlist}  \\
\code{remove\_playlist}  \\
\code{remove\_preference}  \\
\code{replace\_song\_in\_playlist}  \\
\code{replace\_song\_in\_playlist+play\_song}  \\
\code{reset\_playing\_list}  \\
\code{reset\_playlist}  \\
\code{resume\_song}  \\
\code{sort\_playing\_list}  \\
\code{sort\_playlist}  \\
\code{stop\_song}  \\
\code{stop\_song\_after}  \\
\code{vol\_decrease}  \\
\code{vol\_increase}  \\
\code{vol\_max}  \\
\code{vol\_min}  \\
\code{vol\_set}  \\

\end{document}